\documentclass[10pt, a4paper]{article}

\usepackage[]{lrec-coling2024} % this is the new style

\title{Knowledge Graph Representation for Political Information Sources}

\name{Tinatin Osmonova, Alexey Tikhonov, Ivan P. Yamshchikov} 

\address{OSCE Academy, Bishkek\\
Inworld.AI,  Berlin \\
CAIRO, Technical University of Applied Sciences W{\"u}rzburg-Schweinfurt (THWS),  W{\"u}rzburg \\
ivan.yamshchikov@thws.de}

\abstract{
With the rise of computational social science, many scholars utilize data analysis and natural language processing tools to analyze social media, news articles, and other accessible data sources for examining political and social discourse. Particularly, the study of the emergence of echo-chambers due to the dissemination of specific information has become a topic of interest in mixed methods research areas. In this paper, we analyze data collected from two news portals, Breitbart News (BN) and New York Times (NYT) to prove the hypothesis that the formation of echo-chambers can be partially explained on the level of an individual information consumption rather than a collective topology of individuals' social networks. Our research findings are presented through knowledge graphs, utilizing a dataset spanning 11.5 years gathered from BN and NYT media portals. We demonstrate that the application of knowledge representation techniques to the aforementioned news streams highlights, contrary to common assumptions, shows relative "internal" neutrality of both sources and polarizing attitude towards a small fraction of entities. Additionally, we argue that such characteristics in information sources lead to fundamental disparities in audience worldviews, potentially acting as a catalyst for the formation of echo-chambers. 
 \\ \newline }

\begin{document}

\maketitleabstract

\section{Introduction}
A knowledge graph, also known as a semantic network, was initially introduced by C. Hoede and F.N. Stokman as a tool for representing the content of medical and sociological texts \cite{nurdiati200825}.  Constructing increasingly larger graphs with the intent of accumulating knowledge was initially deemed to provide a resultant structure capable of operating as an expert system proficient in investigating causes and computing the consequences of certain decisions.
 
The concept of knowledge graph co-evolved with the rise of computational social science \cite{conte2012manifesto} and digital data analysis methods \cite{rogers2013digital}. Access to open sources on the Internet has facilitated the measurement of the dynamics of political debates \cite{Neuman}. Platforms like Twitter and other microblogging services are widely utilized for studying and modeling social and political discourse \cite{Graham}, \cite{Jungherr} , \cite{wang2018deep}. Contemporary researchers even develop a conceptual framework for predicting the morality underlying political tweets\cite{johnson2018classification}. Moreover, knowledge graphs of fact-checked claims, such as ClaimsKG, have been designed. Such tools facilitate structured queries about truth values, authors, dates, journalistic reviews, and various types of metadata \cite{tchechmedjiev2019claimskg}. 

A significant group of studies, advocate usage of graphs for social, political, and business industry data, stating that “graphs greatly increases the clarity of presentation and makes it easier for a reader to understand the data being used”\cite{kastellec2007using} . Additionally, \cite{Abusalih} explains that knowledge graphs serve as indispensable frameworks that underpin intelligent systems. This is achieved by extracting subtle semantic nuances from textual data sourced from a range of vocabularies and semantic repositories. In the past decade, there has been a notable increase in the examination of political discourse within social content in such a way. The authors discuss in detail the connection between political discussions and the language used in them \cite{Chilton}, \cite{Parker}. Furthermore, the literature examines opinion polarization \cite{banisch2019opinion}, attempts to characterize an intuition of the dynamics of the political debate \cite{yamshchikov2019elephants}, and provides techniques for estimating them \cite{merz2016manifesto}, \cite{subramanian2017joint}, \cite{glavavs2017cross},  \cite{subramanian2018hierarchical} or \cite{rasov2020text}. The extensively employed data sources in studies centered on automated text classification for political discourse analysis involve Manifesto Database \cite{manifesto} and the proceedings of the European Parliament \cite{koehn2005europarl}. 

The challenges arising in contemporary studies on observational and discourse analysis are the quality of data \cite{tweedie1994garbage} and the credibility of data sources. It is crucial to apply statistical measures and tests to quantify the impact of poor data quality and bias on the results \cite{Abusalih}. However, quantifying such effects proves comprehensive in the realm of social sciences due to the numerous indigent properties of social datasets \cite{shah2015big}. One significant challenge is associated with the formation of so-called echo-chambers in social structures, which naturally obstruct the propagation of information, reinforcing disparities across various social strata \cite{goldie2014using}, \cite{colleoni2014echo}, \cite{guo2015bayesian} or \cite{harris2015social}. Addressing the credibility of sources, the phenomenon of fake news draws constant attention from media outlets and researchers. According to \cite{55B}, 55\% of online social network users believe they are accurately informed about recent political updates by the media. Consequently, misleading information and false news have the potential to shape certain beliefs and human behaviors. As a solution, several studies \cite{allcott2017social},  \cite{shu2017fake} or \cite{lazer2018science} analyze and propose methods to enhance the quality of information. Additionally, these studies imply the existence of a certain ground truth that could be universally accepted.

Taking existing knowledge and challenges into account, in this work, we study the issue of news representation from a data analysis perspective. We construct two datasets comprising news articles from "alt-right" and "liberal" news platforms, denoted as Breitbart News (BN) and the New York Times (NYT), spanning 11.5 consecutive years (from 2008 to Fall 2019). We demonstrate that information disparities between these news sources are fundamental regardless of the social structures that encapsulate the readers of the aforementioned outlets.  Upon analyzing the findings, we assert that one has to take into consideration these disparities, since they signify fundamental differences in the foundational data that shapes the perspectives, beliefs, and, ultimately, the behavior of readers. Simply put, even if we had no social media information disparities by various news sources could contribute to echo-chamber formation. 

\section{Data and Methodology}
We have parsed two news sites Breitbart News\footnote{\url {https://www.breitbart.com/}} that could be generally associated with the "alt-right" political views and the New York Times\footnote{\url {https://www.nytimes.com/}} associated with "liberal" political views. The choice of these two media platforms was arbitrary to a certain extent. We parsed all news presented on both platforms in the period from 2008 till the fall of 2019.  Using the texts of the news as input data we built an information extraction pipeline aimed to reconstruct a form of knowledge graph out of the news texts. To do that we have used state of the art open information extraction \cite{Stanovsky2018SupervisedOI} and named entity recognition \cite{Peters2017SemisupervisedST} tools of AllenNLP\footnote{\url {https://allennlp.org}}. The outputs of both models are noisy, so in order to stabilize the resulting signal we came up with the heuristics for substring-matching. We used only ARG0 and ARG1 items of open information extractor and all entities of named entity recognition to extract the most useful objects of the articles. For every entity recognized by both methods, we created a vertex in our knowledge graph. We also applied additional manual 'filtering' of the resulting named entities. The procedure to fix the problems of the different spelling and some artifacts of NER and OIE that crowded the list of entities. Finding longer overlapping substrings with high frequencies we matched longer entities with their shorter "parents". The recognized vertexes were connected with an edge that had an estimate of sentiment and subjectivity calculated with TextBlob\footnote{\url {https://textblob.readthedocs.io/en}}. This naive approach yielded a hypergraph of named entities out of both data sources. The weights of the vertexes corresponded to the number of mentions of a given entity. The edges of the graph had three attributes: frequency, polarity, and subjectivity.
To facilitate further research of news coverage and political discourse we share the gathered data\footnote{\url{https://shorturl.at/ntDOT}}.

\section{Do You Know What I Know?}
In this chapter, we explore the acquired knowledge graphs. In Section 3.1, we present a bird's-eye view of the graph, including key properties, and delve into the most contrasting entities and topics with varying coverage in two sources. Section 3.2 revisits the graphs, highlighting aspects crucial for differences in political discourse.

\begin{figure}[h!]
  \centering
  \includegraphics[width=0.9\linewidth]{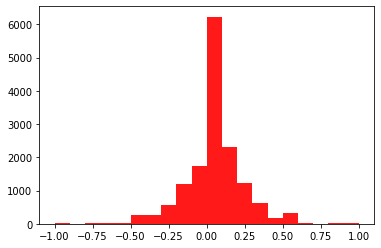}
  \caption{Breitbart News. Distribution of sentiment.}
  \label{fig:bbsent}
\end{figure}
\begin{figure}[h!]
  \centering
  \includegraphics[width=0.9\linewidth]{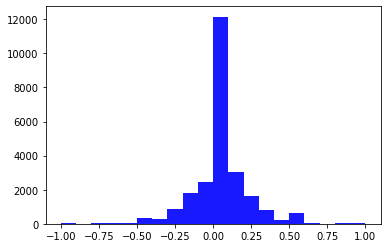}
  \caption{New York Times. Distribution of sentiment.}
  \label{fig:nysent}
\end{figure}
\begin{figure}[h!]
  \centering
  \includegraphics[width=.9\linewidth]{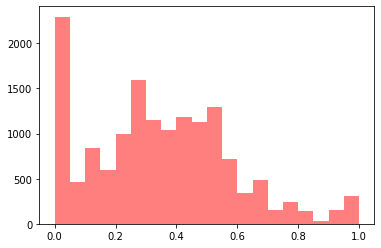}
  \caption{Breitbart News. Distribution of subjectivity.}
  \label{fig:bbsub}
\end{figure}
\begin{figure}[h!]
  \centering
  \includegraphics[width=.9\linewidth]{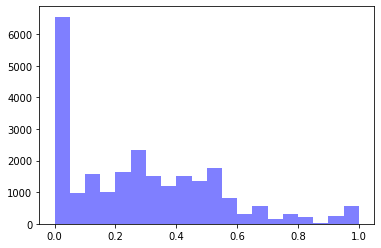}
  \caption{New York Times. Distribution of subjectivity.}
  \label{fig:nysub}
\end{figure}

\subsection{Bird's-eye View} 
\begin{figure*}[h!]
\centering
\includegraphics[width=.80\linewidth]{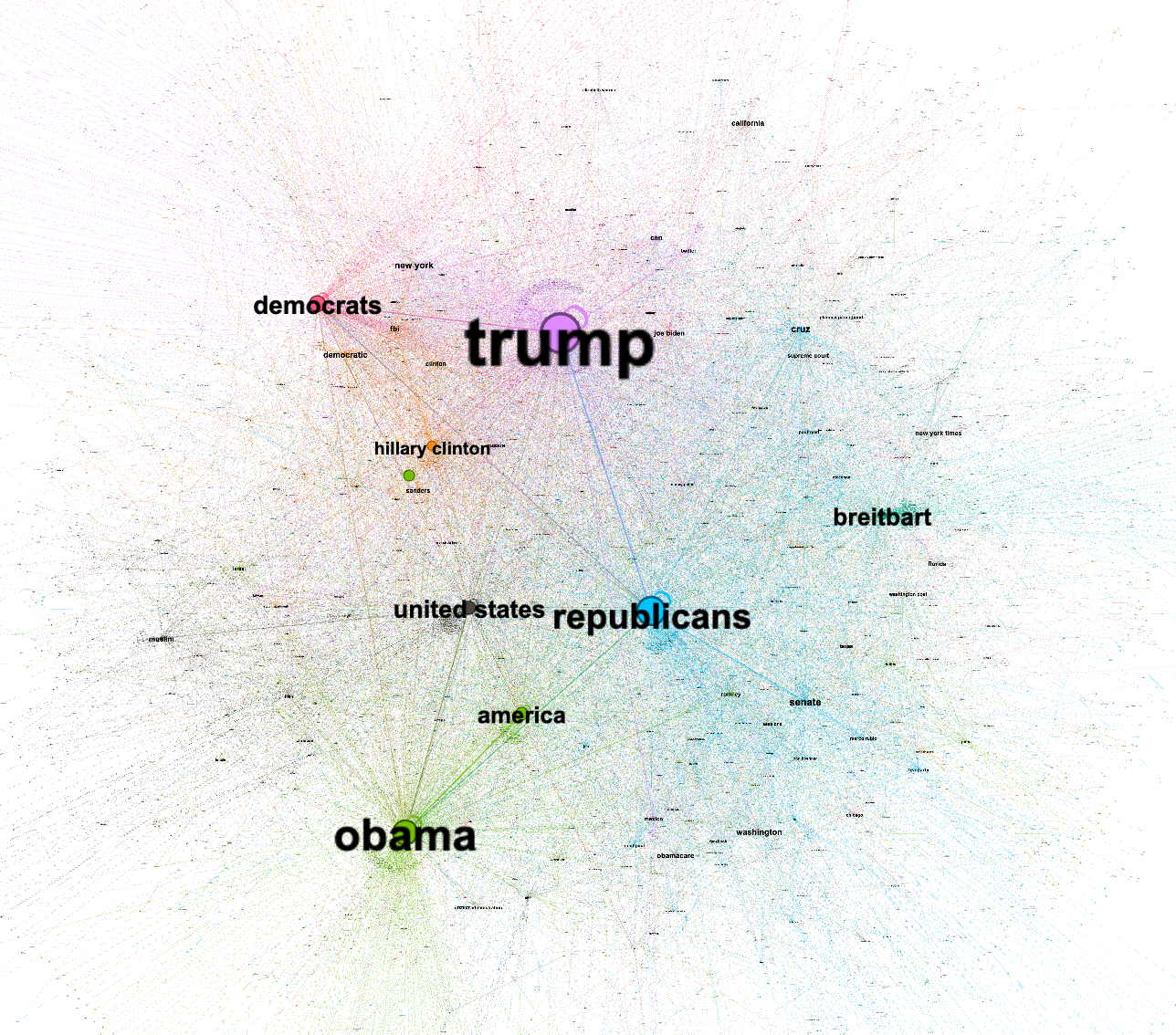}
  \caption{Breitbart News. Overall visualisation of two graphs extracted out of the media sources. The classes found with modularity analysis \cite{blondel2008fast} are highlighted with different colours. Breitbard has smaller number of classes and is centered around US political discourse.}
  \label{fig:sfig1}
  \end{figure*}

\begin{figure*}[h!]
  \centering
  \includegraphics[width=.80\linewidth]{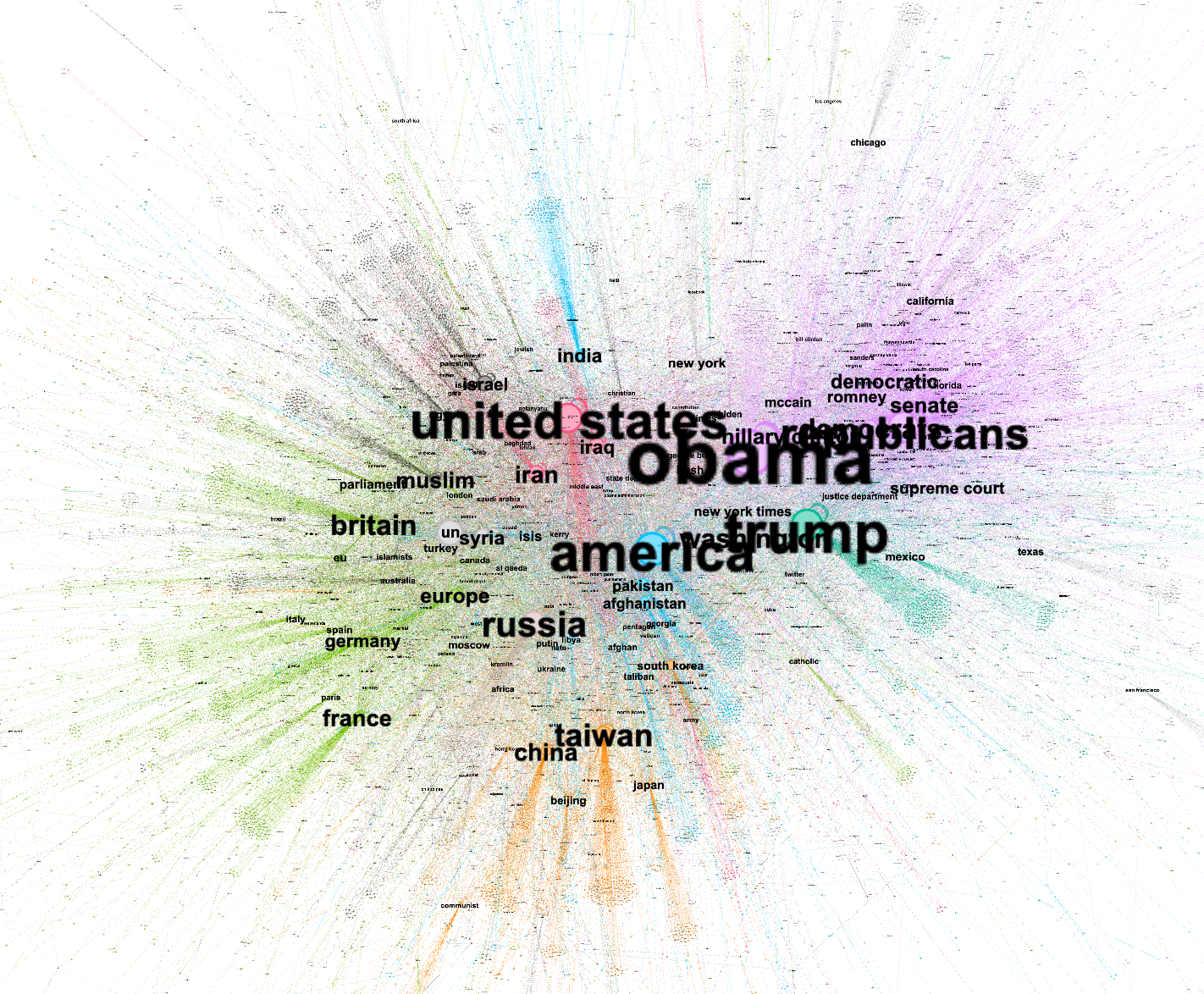}
  \caption{New York Times.Overall visualisation of two graphs extracted out of the media sources. The classes found with modularity analysis \cite{blondel2008fast} are highlighted with different colours. NYT graph has almost twice as many modularity classes and pays way more attention to foreign affairs. This could be partially attributed to the bigger size of the resulting graph, since NYT had more articles published in the studied time period.}
  \label{fig:sfig2}
\end{figure*}

Figures \ref{fig:sfig1} -- \ref{fig:sfig2} show a visualization of two obtained graphs. One can see the divergence of topics: Breitbart is more focused around certain personalities, while the New York Times extensively covers foreign affairs.
Table \ref{tab:accents} shows the first interesting and counter-intuitive result that one can draw when studying obtained graph representations: both media sources are "neutral" on average. Figures \ref{fig:bbsent} -- \ref{fig:nysent} show the distribution of polarity across all edges. The average neutral tone is not a consequence of negatively and positively charged news that balance each other. Distributions in Figures \ref{fig:bbsub} -- \ref{fig:nysub} do not only show that average sentiment across all edges is very close to zero for both graphs, but they also demonstrate and a vast majority of the analyzed relations are presented in a non-polarizing way (at least to the extent to which modern NLP method can distinguish polarity). One can also see the corresponding distributions of subjectivity that are similar for both sources. For the NYT Spearman correlation between polarity and subjectivity is $31\%$, for Breitbart, it is $23\%$.

\begin{table}
\centering
\begin{tabular}{lcrr}
\hline
\textbf{Data} & \textbf{Radius} & \textbf{Diameter} & \textbf{Modularity} \\
\hline
BN            & 6               & 11      & 0.43       \\
NYT       & 7               & 13          & 0.53        \\\hline
\end{tabular}
\begin{tabular}{l}
\textbf{Average}
\end{tabular}
\begin{tabular}{lcrr}
\hline
\textbf{Data} & \textbf{Path length} & \textbf{Polarity} & \textbf{Subjectivity} \\
\hline
BN  &3.76                      & 0.00              & 0.12            \\
NYT & 3.52                      & -0.00             & 0.08.   \\\hline
\end{tabular}
\caption{Various parameters of the obtained graph representations. Both sources are neutral on average with Breitbart being just above and NYT just below zero average polarity. Breitbart tends to be more subjective, yet average subjectivity for both sources is at around 10\%, with NYT a bit more objective.}
\label{tab:accents}
\end{table}


Both media sites try to present themselves to the reader as neutral on average and moderately subjective. This stands to reason: an average reader probably neither wants to feel that she wears rose-tinted glasses nor wants to constantly read that the doom is nigh. Majority of the news are neutral, extremely positive and extremely negative news are rare in both sources. At the same time both sources tend to point bias in the coverage "on the other side".  Another interesting line of thought that could be developed when regarding Table \ref{tab:accents} is the connection between right political actors and propagation of conspiracy theories, see, for example, \cite{hellinger2018conspiracies}. Indeed, the Breitbart graph has smaller modularity and comparable path length. This could imply a lower encapsulation of topics and a higher tendency to connect remote entities. Even a first bird's eye view gives several fundamental insights:
\begin{itemize}
    \item when assessed formally both right and left media demonstrate qualitatively comparable behavior; they try to cover the news in a relatively neutral tone with a pinch of subjectivity;
    \item the coverage of various topics differs significantly; the entities that Brietbart constantly covers tend to be people and actors of domestic US politics, whereas NYT pays more attention to institutions and international affairs;
    \item the overall differences between formally obtained knowledge structures that could proxy right and left world-view are minute, despite our intuition telling us otherwise.
\end{itemize} 
\subsection{Politics of Contrasts}
\begin{figure*}[t!]
    \centering
    \includegraphics[width=\textwidth]{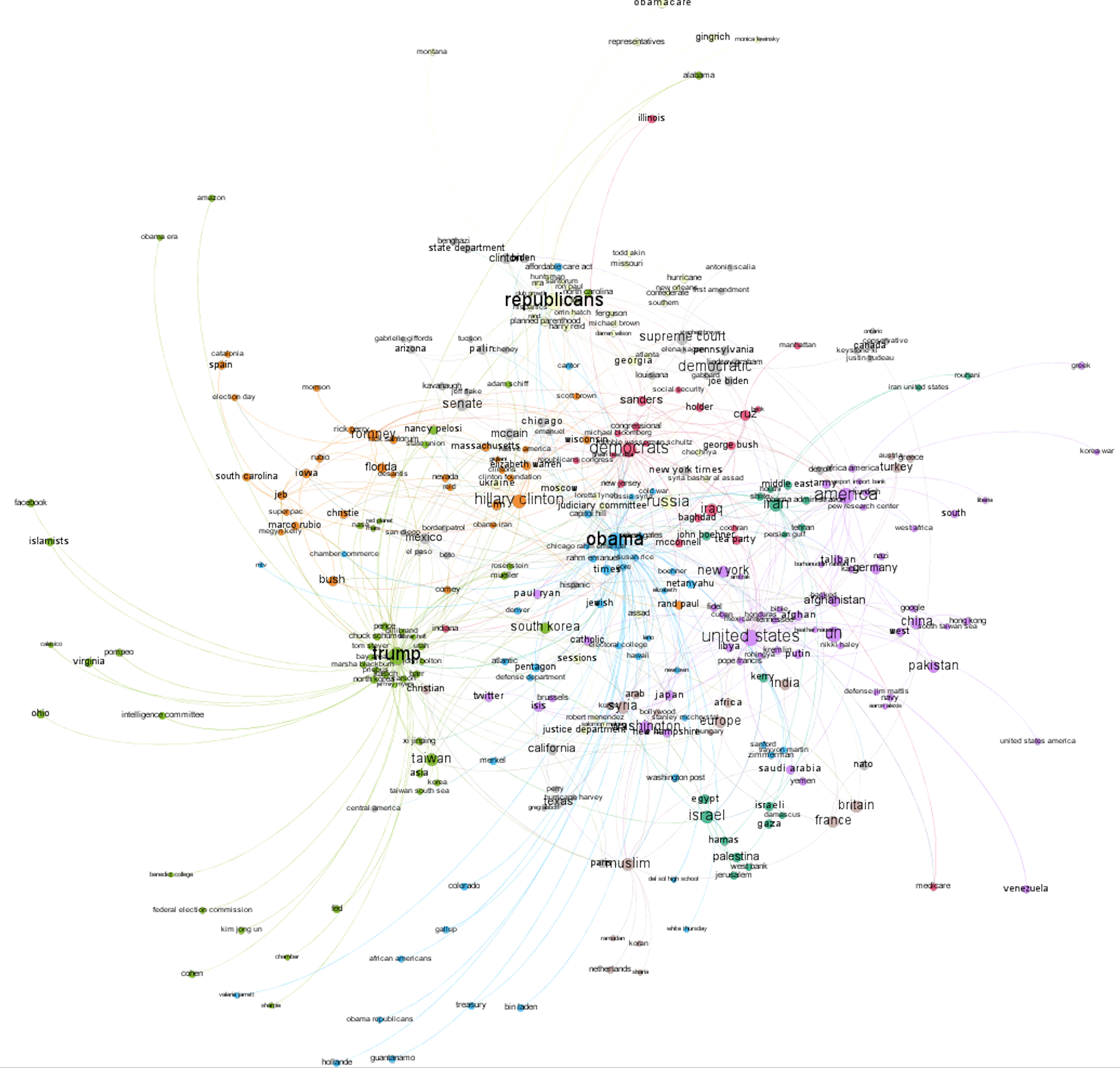}
    \caption{Sub-graph of contrasting edges. These are the edges for which the sign of polarity for BN and NYT is different.}
\label{pic:ed}
\end{figure*}

\begin{figure*}[h!]
    \centering
    \includegraphics[width=0.8\textwidth]{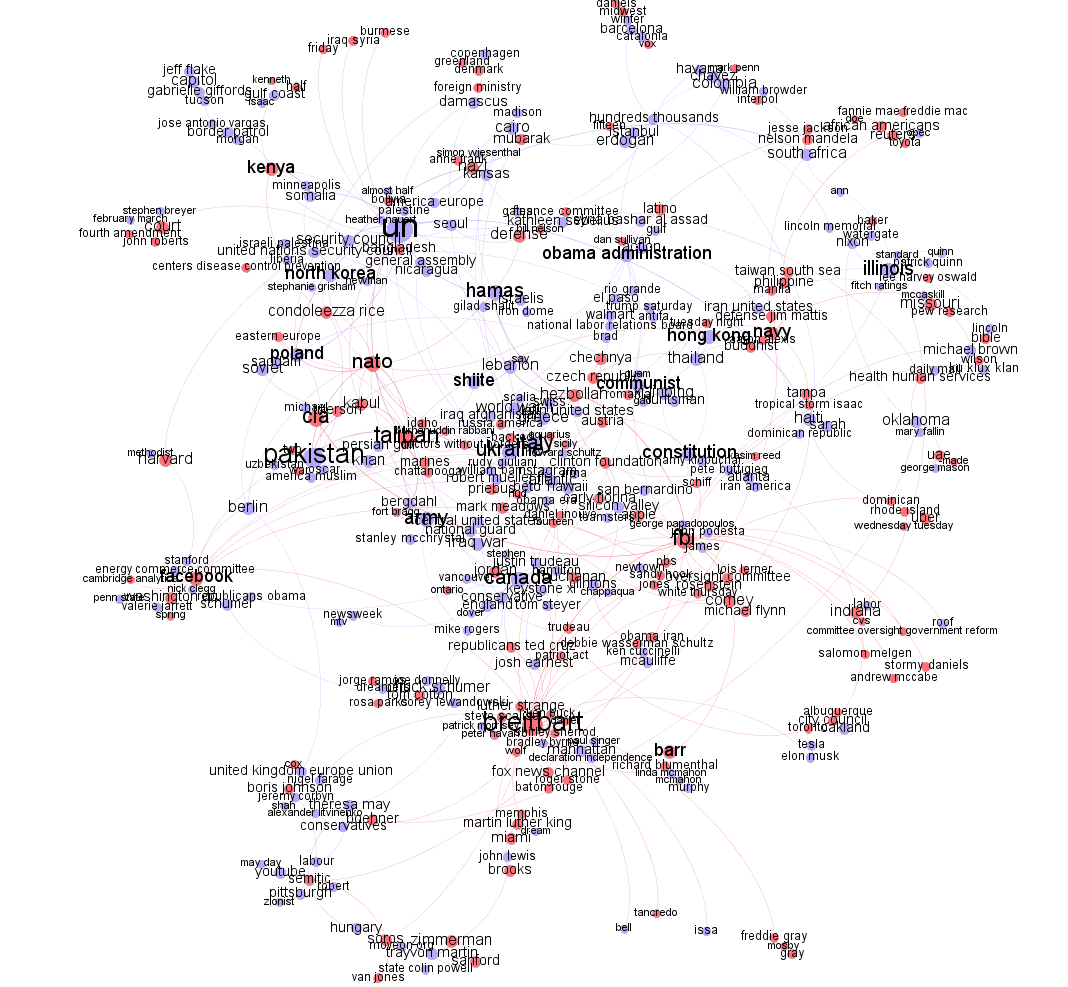}
    \caption{Sub-graph of contrasting vertexes. These are the vertexes for which the average of polarity of the adjacent edges is the highest. Blue nodes are shifted towards NYT, red — towards BN.}
\label{pic:v}
\end{figure*}

Figure \ref{pic:ed} shows a joint graph of the most polarized edges. These are the edges between entities for which the polarity in NYT and Breitbart has a different sign. Similarly, in Figure \ref{pic:v} one could see the most contrasting vertexes. These are the entities that have the highest average polarity of the adjacent edge. Effectively these are the representation of the polarizing topics and are covered with different polarity in both news sources.

An interesting difference between the graph of contrasting edges and the graph of contrasting nodes is that the former is mostly populated with domestic political actors, whereas the latter up to a large extent consists of entities connected with foreign affairs. This is interesting. Certain relationships between entities tend to be more polarizing for domestic issues and local politicians, yet when averaged over several such relationships across time the foreign affairs and institutions come forward. This is the same pattern that we saw earlier. One could speculate that contrasting edges highlight certain local events centered around specific politicians. Such events could be highly polarizing yet temporal. At the same time institutions and global affairs might not be as polarizing as a local scandal, yet the position of both sides on them is persistent, so when averaging across adjacent edges one sees Figure \ref{pic:v}. 

This highlights the fundamental difference between the sources. Though on macro-level both outlets prefer to stick to neutral coverage and refrain from subjectivity when it comes to certain entities and topics they provide different evaluations and tend to be more subjective in these cases. The combination of these two factors is extremely unfortunate since it facilitates social conflict. Indeed, every reader is perfectly convinced that her news source is relevant, objective, and non-biased. This also happens to be true in the vast majority of cases. Yet on a handful of key issues, the media takes a more polarizing and subjective position. Moreover, the local polarizing issues tend to be associated with personalities, while longer, fundamental differences are associated with institutions. This could be attributed to the idea of core political beliefs that could be less polarizing yet may be harder to change in the long run.

\section{Discussion}
One of the key contributions of this paper is an attempt to demonstrate that an echo-chamber is not exactly a phenomenon based solely on the topology of human social networks. Using modern language processing methods and straight-forward knowledge representation we show that two different media sources paint two different pictures of the political reality. Yet these differences are less obvious than we tend to think and are more subtle. Surprisingly low average polarity and subjectivity for both knowledge structures are extremely intriguing. Assuming there is no ill will on the side of the publisher one can try to explain why the overwhelming amount of news articles try to be non-polarizing and non-subjective and with these attempts reinforce the echo-chambers around them without even trying. 
Could echo-chambers be a consequence of human psychological trust mechanisms on top of certain social structure formation? In \cite{levine2014truth} and \cite{clare2019documenting} the authors discuss the truth-default theory. They demonstrate that when people cognitively process the content of others’ communication, they typically do so in a manner characterized by unquestioned, passive acceptance. We could speculate that such behavior naturally transfers to the news sources. High neutrality and low subjectivity reinforce this truth-default. Since the preferred news outlet is often objective and neutral the reader tends to ignore or accept rare polarizing and subjective articles and dismiss the counter-argument of the other side, since in an overwhelming majority of the cases the criticism was not applicable. This might be wild speculation that demands further experimental verification. However, the very idea that echo-chamber formation could be attributed to the personal rather than collective behavior is new to our knowledge.

\section{Conclusion}
In this paper, we present the graphs of entities that correspond to two major "alt-right" and "liberal" news media and their coverage of the mentioned entities and relations between them. The graphs are obtained without any expert knowledge solely with NLP instruments and methods of knowledge representation. Analyzing obtained graphs we show that despite common intuition they exhibit a lot of structural similarities. We also highlight fundamental differences that could be attributed to the formation of echo-chambers and certain biases on the world perception. We suggest that the formation of echo-chambers has more to do with the structure of information consumption and certain core beliefs of the individual rather than social structure that encompasses the aforementioned person. 
\section*{Limitations}
The study covers the period from 2008 to the Fall of 2019, excluding updates beyond 2019. It refrains from a detailed examination of the political aspects and perspectives of Breitbart News and New York Times readers, and it does not develop additional discussions on the global order. Considering recent global crises like wars, economic downturns in specific nations, and the worldwide impact of the COVID-19 pandemic, we anticipate that applying our methodology to recent-year data may produce slightly different findings. Nonetheless, in an effort to encourage transparent research in knowledge representation for social sciences, we provide access to our collected datasets.
\section*{Ethics Statement}
Our work prioritizes transparency and relies on data collected from open sources. We refrain from making political judgments in our discussion notes to prevent discrimination and minimize potential societal harm.

\nocite{*}
\section{Bibliographical References}\label{sec:reference}

\bibliographystyle{lrec-coling2024-natbib}
\bibliography{lrec-coling2024}     

\begin{thebibliography}{82}
\expandafter\ifx\csname natexlab\endcsname\relax\def\natexlab#1{#1}\fi

\bibitem[{Abu-Salih and Beheshti(2021)}]{Abusalih}
Al-Tawil M. Aljarah I. Faris H. Wongthongtham P. Chan K.~Y. Abu-Salih, B. and
  A.~Beheshti. 2021.
\newblock Relational learning analysis of social politics using knowledge graph
  embedding.
\newblock \emph{Data Mining and Knowledge Discovery}, 35(1):1497–1536.

\bibitem[{Aho and Ullman(1972)}]{Aho:72}
Alfred~V. Aho and Jeffrey~D. Ullman. 1972.
\newblock \emph{The Theory of Parsing, Translation and Compiling}, volume~1.
\newblock Prentice-Hall, Englewood Cliffs, NJ.

\bibitem[{Allcott and Gentzkow(2017)}]{allcott2017social}
Hunt Allcott and Matthew Gentzkow. 2017.
\newblock Social media and fake news in the 2016 election.
\newblock \emph{Journal of economic perspectives}, 31(2):211--36.

\bibitem[{{American Psychological Association}(1983)}]{APA:83}
{American Psychological Association}. 1983.
\newblock \emph{Publications Manual}.
\newblock American Psychological Association, Washington, DC.

\bibitem[{Anderson and Auxier(2020)}]{55B}
Monica Anderson and Brooke Auxier. 2020.
\newblock 55% of u.s. social media users say they are ‘worn out’ by
  political posts and discussions.
\newblock In \emph{Pew Research Center}.

\bibitem[{Ando and Zhang(2005)}]{Ando2005}
Rie~Kubota Ando and Tong Zhang. 2005.
\newblock \href {https://www.jmlr.org/papers/volume6/ando05a/ando05a.pdf} {A
  framework for learning predictive structures from multiple tasks and
  unlabeled data}.
\newblock \emph{Journal of Machine Learning Research}, 6:1817--1853.

\bibitem[{Andrew and Gao(2007)}]{andrew2007scalable}
Galen Andrew and Jianfeng Gao. 2007.
\newblock \href {https://dl.acm.org/doi/abs/10.1145/1273496.1273501} {Scalable
  training of {$L_1$}-regularized log-linear models}.
\newblock In \emph{Proceedings of the 24th International Conference on Machine
  Learning}, pages 33--40.

\bibitem[{Banisch and Olbrich(2019)}]{banisch2019opinion}
Sven Banisch and Eckehard Olbrich. 2019.
\newblock Opinion polarization by learning from social feedback.
\newblock \emph{The Journal of Mathematical Sociology}, 43(2):76--103.

\bibitem[{Barbieri et~al.(2020)Barbieri, Camacho-Collados, Neves, and
  Espinosa-Anke}]{barbieri-etal-2020-tweeteval}
Francesco Barbieri, Jose Camacho-Collados, Leonardo Neves, and Luis
  Espinosa-Anke. 2020.
\newblock Tweeteval: Unified benchmark and comparative evaluation for tweet
  classification.
\newblock \emph{arXiv preprint arXiv:2010.12421}.

\bibitem[{Blondel et~al.(2008)Blondel, Guillaume, Lambiotte, and
  Lefebvre}]{blondel2008fast}
Vincent~D Blondel, Jean-Loup Guillaume, Renaud Lambiotte, and Etienne Lefebvre.
  2008.
\newblock Fast unfolding of communities in large networks.
\newblock \emph{Journal of statistical mechanics: theory and experiment},
  2008(10):P10008.

\bibitem[{Bojanowski et~al.(2017)Bojanowski, Grave, Joulin, and
  Mikolov}]{bojanowski2017enriching}
Piotr Bojanowski, Edouard Grave, Armand Joulin, and Tomas Mikolov. 2017.
\newblock Enriching word vectors with subword information.
\newblock \emph{Transactions of the Association for Computational Linguistics},
  5:135--146.

\bibitem[{Borel(1921)}]{borel1921}
E.~Borel. 1921.
\newblock La theorie du jeu et les equations integrales a noyau symetrique.
\newblock \emph{Comptes rendus hebdomadaires des seances de l'Academie des
  sciences}, (173):1304--1308.

\bibitem[{Campigotto et~al.(2014)Campigotto, Conde-Céspedes, and
  Guillaume}]{article4}
Romain Campigotto, Patricia Conde-Céspedes, and Jean-Loup Guillaume. 2014.
\newblock A generalized and adaptive method for community detection.

\bibitem[{Chandra et~al.(1981)Chandra, Kozen, and Stockmeyer}]{Chandra:81}
Ashok~K. Chandra, Dexter~C. Kozen, and Larry~J. Stockmeyer. 1981.
\newblock \href {https://doi.org/10.1145/322234.322243} {Alternation}.
\newblock \emph{Journal of the Association for Computing Machinery},
  28(1):114--133.

\bibitem[{Chen et~al.(2017)Chen, Zhang, Wang, Yang, and Li}]{chen2017opinion}
Wei Chen, Xiao Zhang, Tengjiao Wang, Bishan Yang, and Yi~Li. 2017.
\newblock Opinion-aware knowledge graph for political ideology detection.
\newblock In \emph{IJCAI}, pages 3647--3653.

\bibitem[{Chilton(2004)}]{Chilton}
P.~Chilton. 2004.
\newblock \emph{Analysing political discourse: Theory and practice}.
\newblock Routledge.

\bibitem[{Clare and Levine(2019)}]{clare2019documenting}
David~D Clare and Timothy~R Levine. 2019.
\newblock Documenting the truth-default: The low frequency of spontaneous
  unprompted veracity assessments in deception detection.
\newblock \emph{Human Communication Research}, 45(3):286--308.

\bibitem[{Colleoni et~al.(2014)Colleoni, Rozza, and
  Arvidsson}]{colleoni2014echo}
Elanor Colleoni, Alessandro Rozza, and Adam Arvidsson. 2014.
\newblock Echo chamber or public sphere? predicting political orientation and
  measuring political homophily in twitter using big data.
\newblock \emph{Journal of communication}, 64(2):317--332.

\bibitem[{Conte et~al.(2012)Conte, Gilbert, Bonelli, Cioffi-Revilla, Deffuant,
  Kertesz, Loreto, Moat, Nadal, Sanchez et~al.}]{conte2012manifesto}
Rosaria Conte, Nigel Gilbert, Giulia Bonelli, Claudio Cioffi-Revilla, Guillaume
  Deffuant, Janos Kertesz, Vittorio Loreto, Suzy Moat, J-P Nadal, Anxo Sanchez,
  et~al. 2012.
\newblock Manifesto of computational social science.
\newblock \emph{The European Physical Journal Special Topics}, 214(1):325--346.

\bibitem[{Cooley and Tukey(1965)}]{ct1965}
James~W. Cooley and John~W. Tukey. 1965.
\newblock \href
  {https://www.ams.org/journals/mcom/1965-19-090/S0025-5718-1965-0178586-1/S0025-5718-1965-0178586-1.pdf}
  {An algorithm for the machine calculation of complex {F}ourier series}.
\newblock \emph{Mathematics of Computation}, 19(90):297--301.

\bibitem[{Delobelle et~al.(2019)Delobelle, Cunha, Cano, Peperkamp, and
  Berendt}]{article10}
Pieter Delobelle, Murilo Cunha, Eric Cano, Jeroen Peperkamp, and Bettina
  Berendt. 2019.
\newblock \href {https://doi.org/10.18653/v1/P19-2028} {Computational ad
  hominem detection}.
\newblock pages 203--209.

\bibitem[{Gaisbauer et~al.(2023)Gaisbauer, Pournaki, Banisch, and
  Olbrich}]{gaisbauer2023grounding}
Felix Gaisbauer, Armin Pournaki, Sven Banisch, and Eckehard Olbrich. 2023.
\newblock Grounding force-directed network layouts with latent space models.
\newblock \emph{Journal of Computational Social Science}, pages 1--33.

\bibitem[{Garimella et~al.(2016)Garimella, Morales, Gionis, and
  Mathioudakis}]{article7}
Kiran Garimella, Gianmarco Morales, Aristides Gionis, and Michael Mathioudakis.
  2016.
\newblock \href {https://doi.org/10.1145/2835776.2835792} {Quantifying
  controversy in social media}.
\newblock pages 33--42.

\bibitem[{Glava{\v{s}} et~al.(2017)Glava{\v{s}}, Nanni, and
  Ponzetto}]{glavavs2017cross}
Goran Glava{\v{s}}, Federico Nanni, and Simone~Paolo Ponzetto. 2017.
\newblock Cross-lingual classification of topics in political texts.
\newblock In \emph{Proceedings of the Second Workshop on NLP and Computational
  Social Science}, pages 42--46.

\bibitem[{Goldie et~al.(2014)Goldie, Linick, Jabbar, and
  Lubienski}]{goldie2014using}
David Goldie, Matthew Linick, Huriya Jabbar, and Christopher Lubienski. 2014.
\newblock Using bibliometric and social media analyses to explore the “echo
  chamber” hypothesis.
\newblock \emph{Educational Policy}, 28(2):281--305.

\bibitem[{Graham et~al.(2016)Graham, Jackson, and Broersma}]{Graham}
T.~Graham, D.~Jackson, and M.~Broersma. 2016.
\newblock New platform, old habits? candidates use of twitter during the 2010
  british and dutch general election campaigns.
\newblock \emph{New Media \& Society}, 18(5):765--783.

\bibitem[{Gross and Wagner(1950)}]{grosswagner1950}
O.~Gross and R.~Wagner. 1950.
\newblock A continuous colonel blotto game.
\newblock RAND Research Memorandum.

\bibitem[{Guo et~al.(2015)Guo, Blundell, Wallach, and Heller}]{guo2015bayesian}
Fangjian Guo, Charles Blundell, Hanna Wallach, and Katherine Heller. 2015.
\newblock The bayesian echo chamber: Modeling social influence via linguistic
  accommodation.
\newblock In \emph{Artificial Intelligence and Statistics}, pages 315--323.

\bibitem[{Gusfield(1997)}]{Gusfield:97}
Dan Gusfield. 1997.
\newblock \href
  {https://www.cambridge.org/core/books/algorithms-on-strings-trees-and-sequences/F0B095049C7E6EF5356F0A26686C20D3}
  {\emph{Algorithms on Strings, Trees and Sequences}}.
\newblock Cambridge University Press, Cambridge, UK.

\bibitem[{Harris and Harrigan(2015)}]{harris2015social}
Lisa Harris and Paul Harrigan. 2015.
\newblock Social media in politics: The ultimate voter engagement tool or
  simply an echo chamber?
\newblock \emph{Journal of Political Marketing}, 14(3):251--283.

\bibitem[{Hellinger(2018)}]{hellinger2018conspiracies}
Daniel~C Hellinger. 2018.
\newblock \emph{Conspiracies and conspiracy theories in the age of trump}.
\newblock Springer.

\bibitem[{Jacomy et~al.(2014)Jacomy, Venturini, Heymann, and
  Bastian}]{article5}
Mathieu Jacomy, Tommaso Venturini, Sebastien Heymann, and Mathieu Bastian.
  2014.
\newblock \href {https://doi.org/10.1371/journal.pone.0098679} {Forceatlas2, a
  continuous graph layout algorithm for handy network visualization designed
  for the gephi software}.
\newblock \emph{PloS one}, 9:e98679.

\bibitem[{Jiang et~al.(2021)Jiang, Ren, and Ferrara}]{article8}
Julie Jiang, Xiang Ren, and Emilio Ferrara. 2021.
\newblock \href {https://doi.org/10.2196/29570} {Social media polarization and
  echo chambers: A case study of covid-19}.

\bibitem[{Jin et~al.(2022)Jin, Lalwani, Vaidhya, Shen, Ding, Lyu, Sachan,
  Mihalcea, and Schölkopf}]{article11}
Zhijing Jin, Abhinav Lalwani, Tejas Vaidhya, Xiaoyu Shen, Yiwen Ding, Zhiheng
  Lyu, Mrinmaya Sachan, Rada Mihalcea, and Bernhard Schölkopf. 2022.
\newblock \href {https://doi.org/2202.13758} {Logical fallacy detection}.
\newblock page~1.

\bibitem[{Johnson and Goldwasser(2018)}]{johnson2018classification}
Kristen Johnson and Dan Goldwasser. 2018.
\newblock Classification of moral foundations in microblog political discourse.
\newblock In \emph{Proceedings of the 56th Annual Meeting of the Association
  for Computational Linguistics (Volume 1: Long Papers)}, pages 720--730.

\bibitem[{Jungherr(2014)}]{Jungherr}
A.~Jungherr. 2014.
\newblock The logic of political coverage on twitter: Temporal dynamics and
  content.
\newblock \emph{Journal of Communication}, 64(2):239--259.

\bibitem[{Kastellec and Leoni(2007)}]{kastellec2007using}
Jonathan~P Kastellec and Eduardo~L Leoni. 2007.
\newblock Using graphs instead of tables in political science.
\newblock \emph{Perspectives on politics}, pages 755--771.

\bibitem[{Kim(2014)}]{kim2014}
Y.~Kim. 2014.
\newblock Convolutional neural networks for sentence classification.
\newblock ArXiv:1408.5882.

\bibitem[{Koehn(2005)}]{koehn2005europarl}
Philipp Koehn. 2005.
\newblock Europarl: A parallel corpus for statistical machine translation.
\newblock In \emph{MT summit}, volume~5, pages 79--86. Citeseer.

\bibitem[{Kovenock and Roberson(2015)}]{kovenockroberson2015}
D.~Kovenock and B.~Roberson. 2015.
\newblock Generalizations of the general lotto and colonel blotto games.
\newblock CESifo Working Paper 5291.

\bibitem[{Krause et~al.(2018)Krause, Lehmann, Lewandowski, Matthie{\ss}, Merz,
  Regel, and Werner}]{krause2018manifesto}
Werner Krause, Pola Lehmann, Jirka Lewandowski, Theres Matthie{\ss}, Nicolas
  Merz, Sven Regel, and Annika Werner. 2018.
\newblock Manifesto corpus.
\newblock \emph{WZB Berlin Social Science Center}.

\bibitem[{Lacewell and Werner(2013)}]{lacewell2013coder}
Onawa~P Lacewell and Annika Werner. 2013.
\newblock Coder training: Key to enhancing coding reliability and estimate
  validity.

\bibitem[{Lan et~al.(2008)Lan, Tan, Su, and Lu}]{lan2008supervised}
Man Lan, Chew~Lim Tan, Jian Su, and Yue Lu. 2008.
\newblock Supervised and traditional term weighting methods for automatic text
  categorization.
\newblock \emph{IEEE transactions on pattern analysis and machine
  intelligence}, 31(4):721--735.

\bibitem[{Laslier and Picard(2002)}]{laslierpicard2002}
J.-F. Laslier and N.~Picard. 2002.
\newblock Distributive politics and electoral competition.
\newblock \emph{Journal of Economic Theory}, (103):106--130.

\bibitem[{Lazer et~al.(2018)Lazer, Baum, Benkler, Berinsky, Greenhill, Menczer,
  Metzger, Nyhan, Pennycook, Rothschild et~al.}]{lazer2018science}
David~MJ Lazer, Matthew~A Baum, Yochai Benkler, Adam~J Berinsky, Kelly~M
  Greenhill, Filippo Menczer, Miriam~J Metzger, Brendan Nyhan, Gordon
  Pennycook, David Rothschild, et~al. 2018.
\newblock The science of fake news.
\newblock \emph{Science}, 359(6380):1094--1096.

\bibitem[{Lehmann et~al.(2017)Lehmann, Matthie{\ss}, Merz, Regel, and
  Werner}]{manifesto}
P.~Lehmann, T.~Matthie{\ss}, N.~Merz, S.~Regel, and A.~Werner. 2017.
\newblock Manifesto corpus 2017-1.
\newblock WZB Berlin Social Science Center.

\bibitem[{Levine(2014)}]{levine2014truth}
Timothy~R Levine. 2014.
\newblock Truth-default theory (tdt) a theory of human deception and deception
  detection.
\newblock \emph{Journal of Language and Social Psychology}, 33(4):378--392.

\bibitem[{Loria et~al.(2018)}]{loria2018textblob}
Steven Loria et~al. 2018.
\newblock textblob documentation.
\newblock \emph{Release 0.15}, 2(8):269.

\bibitem[{Martin et~al.(2011)Martin, Brown, Klavans, and Boyack}]{article6}
Shawn Martin, W.~Brown, Richard Klavans, and Kevin Boyack. 2011.
\newblock \href {https://doi.org/10.1117/12.871402} {Openord: An open-source
  toolbox for large graph layout}.
\newblock \emph{Proc SPIE}, 7868:786806.

\bibitem[{Merz et~al.(2016)Merz, Regel, and Lewandowski}]{merz2016manifesto}
Nicolas Merz, Sven Regel, and Jirka Lewandowski. 2016.
\newblock The manifesto corpus: A new resource for research on political
  parties and quantitative text analysis.
\newblock \emph{Research \& Politics}, 3(2):2053168016643346.

\bibitem[{Mikhaylov et~al.(2012)Mikhaylov, Laver, and
  Benoit}]{mikhaylov2012coder}
Slava Mikhaylov, Michael Laver, and Kenneth~R Benoit. 2012.
\newblock Coder reliability and misclassification in the human coding of party
  manifestos.
\newblock \emph{Political Analysis}, 20(1):78--91.

\bibitem[{Myerson(1993)}]{myerson1993}
R.B. Myerson. 1993.
\newblock Incentives to cultivate minorities under alternative electoral
  systems.
\newblock \emph{American Political Science Review}, 87:856--869.

\bibitem[{Naseem et~al.(2020)Naseem, Razzak, Musial, and Imran}]{article9}
Usman Naseem, Imran Razzak, Katarzyna Musial, and Muhammad Imran. 2020.
\newblock \href {https://doi.org/10.1016/j.future.2020.06.050} {Transformer
  based deep intelligent contextual embedding for twitter sentiment analysis}.
\newblock \emph{Future Generation Computer Systems}, 113.

\bibitem[{Neuman et~al.(2014)Neuman, Guggenheim, Jang, and Bae}]{Neuman}
R.~Neuman, L.~Guggenheim, S.~Mo Jang, and S.Y. Bae. 2014.
\newblock The dynamics of public attention: Agenda setting theory meets big
  data.
\newblock \emph{Journal of Communication}, 64(2):193--214.

\bibitem[{Nurdiati and Hoede(2008)}]{nurdiati200825}
Sri Nurdiati and Cornelis Hoede. 2008.
\newblock 25 years development of knowledge graph theory: the results and the
  challenge.
\newblock \emph{Memorandum}, 1876(2):1--10.

\bibitem[{Osorio(2013)}]{osorio2013}
A.~Osorio. 2013.
\newblock The lottery blotto game.
\newblock \emph{Economics Letters}, 120(2):164--166.

\bibitem[{Parker(2014)}]{Parker}
I.~Parker. 2014.
\newblock \emph{Discourse Dynamics (Psychology Revivals): Critical Analysis for
  Social and Individual Psychology}.
\newblock Routledge.

\bibitem[{Peixoto(2020)}]{peixoto2020latent}
Tiago~P Peixoto. 2020.
\newblock Latent poisson models for networks with heterogeneous density.
\newblock \emph{Physical Review E}, 102(1):012309.

\bibitem[{Peters et~al.(2017)Peters, Ammar, Bhagavatula, and
  Power}]{Peters2017SemisupervisedST}
Matthew~E. Peters, Waleed Ammar, Chandra Bhagavatula, and Russell Power. 2017.
\newblock Semi-supervised sequence tagging with bidirectional language models.
\newblock \emph{ArXiv}, abs/1705.00108.

\bibitem[{Petit et~al.(2020)Petit, Li, and Ali}]{article1}
John Petit, Cong Li, and Khudejah Ali. 2020.
\newblock \href {https://doi.org/10.1016/j.tele.2020.101471} {Fewer people,
  more flames: How pre-existing beliefs and volume of negative comments impact
  online news readers’ verbal aggression}.
\newblock \emph{Telematics and Informatics}, 56:101471.

\bibitem[{Rasooli and Tetreault(2015)}]{rasooli-tetrault-2015}
Mohammad~Sadegh Rasooli and Joel~R. Tetreault. 2015.
\newblock \href {http://arxiv.org/abs/1503.06733} {Yara parser: {A} fast and
  accurate dependency parser}.
\newblock \emph{Computing Research Repository}, arXiv:1503.06733.
\newblock Version 2.

\bibitem[{Rasov et~al.(2020)Rasov, Obabkov, Olbrich, and
  Yamshchikov}]{rasov2020text}
Arsenii Rasov, Ilya Obabkov, Eckehard Olbrich, and Ivan~P Yamshchikov. 2020.
\newblock Text classification for monolingual political manifestos with words
  out of vocabulary.
\newblock In \emph{COMPLEXIS}, pages 149--154.

\bibitem[{Roberson(2006{\natexlab{a}})}]{roberson2006}
B.~Roberson. 2006{\natexlab{a}}.
\newblock The colonel blotto game.
\newblock \emph{Economic Theory}, 29(1):1--24.

\bibitem[{Roberson(2006{\natexlab{b}})}]{roberson}
B.~Roberson. 2006{\natexlab{b}}.
\newblock Pork-barrel politics, discriminatory policies and fiscal federalism.
\newblock Social Science Research Center Berlin (WZB).

\bibitem[{Rogers(2013)}]{rogers2013digital}
Richard Rogers. 2013.
\newblock \emph{Digital methods}.
\newblock MIT press.

\bibitem[{Rossi et~al.(2018)Rossi, Polderman, and Frasca}]{article3}
Wilbert~Samuel Rossi, Jan Polderman, and Paolo Frasca. 2018.
\newblock \href {https://doi.org/10.1109/TCNS.2021.3105616} {The closed loop
  between opinion formation and personalised recommendations}.

\bibitem[{Rosvall and Bergstrom(2008)}]{rosvall2008maps}
Martin Rosvall and Carl~T Bergstrom. 2008.
\newblock Maps of random walks on complex networks reveal community structure.
\newblock \emph{Proceedings of the national academy of sciences},
  105(4):1118--1123.

\bibitem[{Shah et~al.(2015)Shah, Cappella, and Neuman}]{shah2015big}
Dhavan~V Shah, Joseph~N Cappella, and W~Russell Neuman. 2015.
\newblock Big data, digital media, and computational social science:
  Possibilities and perils.
\newblock \emph{The ANNALS of the American Academy of Political and Social
  Science}, 659(1):6--13.

\bibitem[{Shore et~al.(2018)Shore, Baek, and Dellarocas}]{article2}
Jesse Shore, Jiye Baek, and Chrysanthos Dellarocas. 2018.
\newblock \href {https://doi.org/10.25300/MISQ/2018/14558} {Network structure
  and patterns of information diversity on twitter}.
\newblock \emph{MIS Quarterly}, 42.

\bibitem[{Shu et~al.(2017)Shu, Sliva, Wang, Tang, and Liu}]{shu2017fake}
Kai Shu, Amy Sliva, Suhang Wang, Jiliang Tang, and Huan Liu. 2017.
\newblock Fake news detection on social media: A data mining perspective.
\newblock \emph{ACM SIGKDD explorations newsletter}, 19(1):22--36.

\bibitem[{Stanovsky et~al.(2018)Stanovsky, Michael, Zettlemoyer, and
  Dagan}]{Stanovsky2018SupervisedOI}
Gabriel Stanovsky, Julian Michael, Luke Zettlemoyer, and Ido Dagan. 2018.
\newblock Supervised open information extraction.
\newblock In \emph{NAACL-HLT}.

\bibitem[{Subramanian et~al.(2018)Subramanian, Cohn, and
  Baldwin}]{subramanian2018hierarchical}
Shivashankar Subramanian, Trevor Cohn, and Timothy Baldwin. 2018.
\newblock Hierarchical structured model for fine-to-coarse manifesto text
  analysis.
\newblock In \emph{Proceedings of the 2018 Conference of the North American
  Chapter of the Association for Computational Linguistics: Human Language
  Technologies, Volume 1 (Long Papers)}, pages 1964--1974.

\bibitem[{Subramanian et~al.(2017)Subramanian, Cohn, Baldwin, and
  Brooke}]{subramanian2017joint}
Shivashankar Subramanian, Trevor Cohn, Timothy Baldwin, and Julian Brooke.
  2017.
\newblock Joint sentence-document model for manifesto text analysis.
\newblock In \emph{Proceedings of the Australasian Language Technology
  Association Workshop 2017}, pages 25--33.

\bibitem[{Tchechmedjiev et~al.(2019)Tchechmedjiev, Fafalios, Boland, Gasquet,
  Zloch, Zapilko, Dietze, and Todorov}]{tchechmedjiev2019claimskg}
Andon Tchechmedjiev, Pavlos Fafalios, Katarina Boland, Malo Gasquet,
  Matth{\"a}us Zloch, Benjamin Zapilko, Stefan Dietze, and Konstantin Todorov.
  2019.
\newblock Claimskg: a knowledge graph of fact-checked claims.
\newblock In \emph{International Semantic Web Conference}, pages 309--324.
  Springer.

\bibitem[{Thomas et~al.(2006)Thomas, Pang, and Lee}]{thomas2006get}
Matt Thomas, Bo~Pang, and Lillian Lee. 2006.
\newblock Get out the vote: Determining support or opposition from
  congressional floor-debate transcripts.
\newblock \emph{arXiv preprint cs/0607062}.

\bibitem[{Tweedie et~al.(1994)Tweedie, Mengersen, and
  Eccleston}]{tweedie1994garbage}
Richard~L Tweedie, Kerrie~L Mengersen, and John~A Eccleston. 1994.
\newblock Garbage in, garbage out: can statisticians quantify the effects of
  poor data?
\newblock \emph{Chance}, 7(2):20--27.

\bibitem[{Verberne et~al.(2014)Verberne, D’hondt, van~den Bosch, and
  Marx}]{verberne2014automatic}
Suzan Verberne, Eva D’hondt, Antal van~den Bosch, and Maarten Marx. 2014.
\newblock Automatic thematic classification of election manifestos.
\newblock \emph{Information Processing \& Management}, 50(4):554--567.

\bibitem[{Wang et~al.(2018)Wang, Chen, Ren, Yu, Cheng, and Lin}]{wang2018deep}
Zhouxia Wang, Tianshui Chen, Jimmy Ren, Weihao Yu, Hui Cheng, and Liang Lin.
  2018.
\newblock Deep reasoning with knowledge graph for social relationship
  understanding.
\newblock In \emph{Proceedings of the 27th International Joint Conference on
  Artificial Intelligence}, pages 1021--1028.

\bibitem[{Ward et~al.(2011)Ward, Stovel, and Sacks}]{ward2011network}
Michael~D Ward, Katherine Stovel, and Audrey Sacks. 2011.
\newblock Network analysis and political science.
\newblock \emph{Annual Review of Political Science}, 14:245--264.

\bibitem[{Washburn(2013)}]{washburn}
A.~Washburn. 2013.
\newblock Blotto politics.
\newblock \emph{Operations Research}, 61(3):532--543.

\bibitem[{Yamshchikov and Rezagholi(2019)}]{yamshchikov2019elephants}
Ivan~P Yamshchikov and Sharwin Rezagholi. 2019.
\newblock Elephants, donkeys, and colonel blotto.
\newblock In \emph{COMPLEXIS}, pages 113--119.

\bibitem[{Zirn et~al.(2016)Zirn, Glava{\v{s}}, Nanni, Eichorts, and
  Stuckenschmidt}]{zirn2016classifying}
C{\"a}cilia Zirn, Goran Glava{\v{s}}, Federico Nanni, Jason Eichorts, and
  Heiner Stuckenschmidt. 2016.
\newblock Classifying topics and detecting topic shifts in political
  manifestos.

\end{thebibliography}
\end{document}